# Study on Sparse Representation based Classification for Biometric Verification

Zengxi Huang[1], Yiguang Liu[2], Xiaoming Wang[1], Jinrong Hu[1]

[1]School of Computer and Software Engineering, Xihua University, Chengdu, 610039, People's Republic of China.

[2]College of Computer Science, Sichuan University, Chengdu, 610064, People's Republic of China.

**Abstract** In this paper, we propose a multimodal verification system integrating face and ear based on sparse representation based classification (SRC). The face and ear query samples are first encoded separately to derive sparsity-based match scores, and which are then combined with sum-rule fusion for verification. Apart from validating the encouraging performance of SRC-based multimodal verification, this paper also dedicates to provide a clear understanding about the characteristics of SRC-based biometric verification. To this end, two sparsity-based metrics, i.e. spare coding error (SCE) and sparse contribution rate (SCR), are involved, together with face and ear unimodal SRC-based verification. As for the issue that SRC-based biometric verification may suffer from heavy computational burden and verification accuracy degradation with increase of enrolled subjects, we argue that it could be properly resolved by exploiting small random dictionary for sparsity-based score computation, which consists of training samples from a limited number of randomly selected subjects. Experimental results demonstrate the superiority of SRC-based multimodal verification compared to the state-of-the-art multimodal methods like likelihood ratio (LLR), support vector machine (SVM), and the sum-rule fusion methods using cosine similarity, meanwhile the idea of using small random dictionary is feasible in both effectiveness and efficiency.

**Keywords** Multimodal verification; Face and ear; Sparse representation based classification (SRC); Sparsity-based metric; Small random dictionary.

## 1. Introduction

Most recently, the new emerging sparse representation based classification (SRC) techniques have enjoyed great success in biometric identification area. In the literature [1-4], SRC and its extensions have been ascertained to get significant improvement over conventional classifiers like nearest neighbor (NN), nearest feature line (NFL) and nearest subspace (NS) [5-6] in both recognition accuracy and robustness. SRC conducts one-to-many comparison in a unique coding procedure and is naturally exploited for biometric identification. Note that theoretically all classification techniques presented for identification could be used in verification task [7]. However, to the best of our knowledge, until now there has been very little effort on research of SRC for biometric verification.

SRC was first extended to the speaker verification task in [8], where the verification decision is simply based on that whether any training sample of the claimed class gets the biggest coefficient in the coding vector. With similar idea, Li et al. [9] created the dictionary using the total variability i-vectors and presented three sparsity-based metrics for speaker verification. They achieved results almost comparable to the up-to-date approaches. Shin et al. [10] used the sparse coding error (or residual, SCE) as metric and combined the residuals of different color channels of face image with sum-rule. Their approach achieved an equal error rates (EER) of 1.89% and 2.79% on two color face databases, respectively, and outperforms four common methods at least 10%. The performance is very impressive, but such a system based on face color may be sensitive to facial expression, makeup, and illumination changes in practice. Xin et al. [11] utilized the same SRC-based verification in finger vein and got an EER of 0.017% on a database with 600 fingers, which also excels many existing methods. We would like to emphasize that some biometric verification approaches presented recently employ $L_1$-norm technique for feature extraction, and then perform verification in a conventional one-to-one comparison way [12-14]. This category of biometric technologies is different from SRC-based verification and is out of our scope of study.

As a new one-to-many comparison scheme, SRC-based verification is apparently different from common verification approaches. It could possibly have some characteristics or even some limitations needed to be unrevealed or resolved. However, little related research can be found in the previous papers [8-11]. By incorporating non-target classes, SRC-based verification provides a craft competition mechanism to allocate match score to each class by using $L_1$-norm optimization. As such, sparsity-based match scores would be more meaningful for verification than other metrics. However, considering its computational complexity relevant to dictionary scale, SRC-based verification may be very time-consuming, or even prohibitive on large-scale databases, while its verification accuracy may reduce as well. Besides, with the sparsity constraint, SRC would generally only allow very few classes to get a small SCE, while the remainder classes would get distinctly large SCEs. Hence, if the class claimed with valid test sample fails to get a small SCE in the competitive sparse coding, the SRC-based verification is likely to generate a negative response. Moreover, even if the SRC-based verification gets a very low equal error rate (EER), there might have many genuine scores locate near the distribution center of imposter scores, as shown in Fig. 1, thus it is impractical to accept them by tuning operating threshold. That is to say, the verification system may confront a bottleneck in false rejected rate (FRR). Therefore, SRC-based verification may not be suitable for some biometric traits or in some application scenarios.

Driven by above questions, together with the encouraging performance of SRC-based multimodal identification using face and ear in our previous works [15-17], in this paper we study the SRC with two sparsity-based metrics, i.e. SCE and sparse contribution rate (SCR), on biometric verification using face, ear, and their multimodal combination. The SRC-based multimodal system first sparsely encodes the face and ear query samples independently to derive sparsity-based match scores, and then combines them with sum-rule fusion. Moreover, for settling SRC's limitations in verification effectiveness and computational efficiency in large-scale application scenarios, we propose to exploit small random dictionary that consists of training samples from a small part of subjects randomly selected for verification.

In our experiments, both SRC-based multimodal verification methods using SCE or SCR are found significantly superior to the state-of-the-art likelihood ratio (LLR), support vector machine (SVM), and the standard sum-rule fusion methods using cosine similarity as decision metric. Particularly, the multimodal verification with SCE can get an EER of 0.146% on a multimodal database with 6083 testing samples. The face and ear unimodal SRC-based verification approaches also get very impressive results, which are even much better than the three competing multimodal methods. Still, they are not only inferior to SRC-based multimodal methods but also confront evident FRR bottleneck. Based on the experimental results of these SRC-based methods, we confirm the strong correlation between the SRC-based verification performance and the inter-class separability within dictionary. This characteristic could be seen as a selection guideline of SRC for verification applications based on the extensive SRC-based identification reports in literature. The experiments also demonstrate that small random dictionaries can generally bring about comparable or even better verification accuracy than the dictionary with all training samples, and meanwhile use much less time. This implies that using small random dictionary is a feasible solution to settle SRC-based verification's limitations in verification effectiveness and computational efficiency on large-scale databases.

The rest of this paper is organized as follows. In Section 2, we first present two sparsity-based metrics (i.e. SCE and SCR) and the SRC-based multimodal verification system; Secondly, after revealing the limitations of SRC-based verification, we propose a solution of utilizing small random dictionary; Lastly, the SRC-based verification extension combining rank information is presented. Section 3 pays a lot effort on experimental validations and discussions. Finally, we draw conclusions in Section 4.

## 2. SRC-based biometric verification

### 2.1 Two sparsity-based metrics: SCE and SCR

The basic idea of SRC is motivated by the following observation: Given sufficient training samples of each class, a valid test sample could be faithfully represented by the linear combination of entire training sample set, and meanwhile which is generally dominated by training samples of the class associated to the test sample. Suppose that there are $c$ known pattern classes. For convenience, we assume each class has $l$ training samples. Let $A_i = [a_{i,1}, a_{i,2}, \cdots, a_{i,l}] \in R^{N \times l}$ be a matrix formed by the training samples (or feature vectors) of Class $i$, then $A = [A_1, A_2, \cdots, A_c] \in R^{N \times lc}$ is a dictionary composed of all training samples. Given a test sample $y$, it can be represented by $y = A\alpha$, where $\alpha \in R^{lc}$ is the representation coefficient vector. This linear equation system is underdetermined if $N < lc$, then generally its sparsest solution can be sought by solving the following optimization problem [1]:

$$\hat{\alpha} = \arg\min \|\alpha\|_1 \quad \text{s.t.} \quad \|y - A\alpha\|_2 < \varepsilon, \tag{1}$$

where $\|\cdot\|_1$ denotes the $L_1$-norm, which counts the number of nonzero entries in a vector, and $\varepsilon > 0$ is a constant.

Since $\hat{\boldsymbol{\alpha}}$ is achieved, using only the coefficients associated with the $i^{th}$ class, the SCE valude is calculated by

$$\text{SCE:} \quad \eta_i(\boldsymbol{y}) = \|\boldsymbol{y} - A_i \delta_i(\hat{\boldsymbol{\alpha}})\|_2, \tag{2}$$

where $\delta_i: R^N \to R^N$ is the characteristic function that selects the coefficients associated with the Class $i$. The SRC decision rule is: if $\eta_k(\boldsymbol{y}) = \min_i \eta_i(\boldsymbol{y})$, the test sample $\boldsymbol{y}$ is assigned to Class $k$.

SRC and most of its extensions identify a test sample based on comparing the SCEs of all classes. Their superior classification performance in comparison to conventional methods demonstrates that SCE qualifies for measuring the correlation between a test sample and a class, as a distance score. Thus, it is reasonable to use SCE for verification. Considering the binary classification in verification decision, the output is either *genuine* or *imposter* (corresponding to "accept" and "reject", respectively), which can be denoted with 1 and 0, respectively. Given a decision threshold $\theta_{sce}$, the verification rule with SCE can be written as

$$d(\eta_i(\boldsymbol{y})) = \begin{cases} 1 & \text{if } \eta_i(\boldsymbol{y}) \leq \theta_{sce} \\ 0 & \text{if } \eta_i(\boldsymbol{y}) > \theta_{sce} \end{cases}. \tag{3}$$

Wright et al. [1] have presented a metric called sparse concentration index (SCI) to reject outlier, the subject who has no training samples in dictionary, in face identification. Essentially, as described in the Eq. (14) in [1], SCI value depends on the class who contributes the most in sparse coding. Assuming the test sample is not from outlier, generally it should belong to the class with the maximum sparse contribution rate (SCR), as defined in Eq. (4). The larger value of SCR obtained by a class indicates the higher possibility of the test sample belonging to this class. In light of this, SCR could be used as a verification decision metric, a similarity score.

$$\text{SCR:} \quad \rho_i(\hat{\boldsymbol{\alpha}}) = \|\delta_i(\hat{\boldsymbol{\alpha}})\|_1 / \|\hat{\boldsymbol{\alpha}}\|_1 \quad (\delta_i(\hat{\boldsymbol{\alpha}}) \in [0,1]). \tag{4}$$

The verification rule with a given threshold, $\theta_{scr}$, can be written as

$$d(\rho_i(\hat{\boldsymbol{\alpha}})) = \begin{cases} 0 & \text{if } \rho_i(\hat{\boldsymbol{\alpha}}) \leq \theta_{scr} \\ 1 & \text{if } \rho_i(\hat{\boldsymbol{\alpha}}) > \theta_{scr} \end{cases}. \tag{5}$$

Fig. 1 plots the distributions of SCE and SCR scores obtained on GT face database [18]. As for SCE, most the genuine scores distribute in [0, 0.5], while the imposter distribution center is 1.0. On the contrary, almost all imposter scores of SCR are close to 0, there is a bigger overlap between genuine and imposter classes in comparison to that of SCE. This means that the verification based on SCE would be better, which will be validated by the experiments later. The disadvantage of SCR could possibly originate from that Eq. (1) is solved by choosing $\boldsymbol{\alpha}$ to minimize the overall SCE but not the SCR.

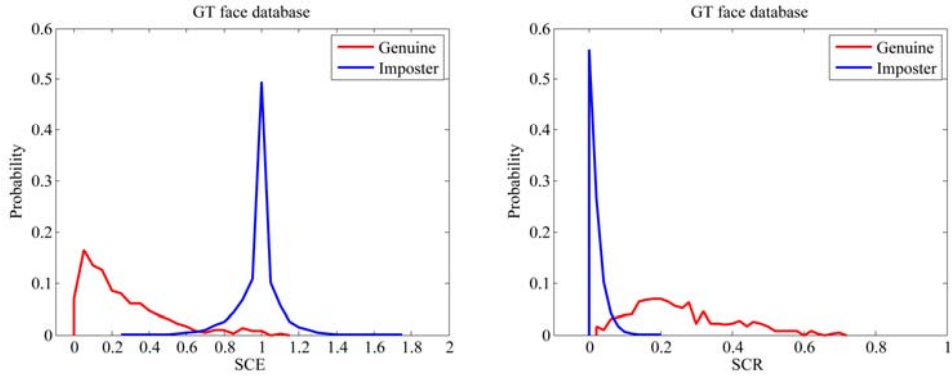

Fig. 1. Distributions of SCE and SCR scores on GT face database.

In an identity recognition process, SRC implicitly compares the test sample with all classes simultaneously through a sparse coding procedure, thereby generating SCE or SCR scores for them. If the system works in verification mode, it will only compare the match score obtained by the claimed class with a given threshold and make a "accept" or "reject" decision. In the identification operation mode, the system will examine all the SCE scores (sorting them in an ascending order, if necessary) and assign the test sample to the class with the least SCE. Therefore, from a viewpoint of operation process, the difference between SRC-based verification and typical SRC identification is very small. Both of them rely on SRC's discriminative capability.

**2.2 SRC-based multimodal verification**

Face recognition is arguably the most extensively studied branch in image-based recognition fields in last decade [19]. Apart from the fundamental challenges it poses, research in face recognition is mainly motivated by its wider range of practical applications relative to other biometric technologies like fingerprint and iris, owing to its advantages: It is natural, non-intrusive, and easy to use [19]. Ear biometric is also non-intrusive and easy to be captured by common camera, and meanwhile has several appealing advantages over the face: the ear has a stable structure with rich information, nearly unaffected by aging and facial expressions [15-17, 20-21]. Although it may be more likely for hair to obscure the ear than the face, automatic ear detection and verification is feasible because user cooperation is generally available in verification scenario [20]. The face and the ear can provide complementary information to each other such that promote the biometric recognition accuracy and robustness. Our previous studies [15-17] have already shown encouraging identification performance of their combination based on SRC-based techniques. Inspired by these works, and to explore the SRC-based verification, in this paper we will use the face, ear, and their multimodal combination as three biometric examples.

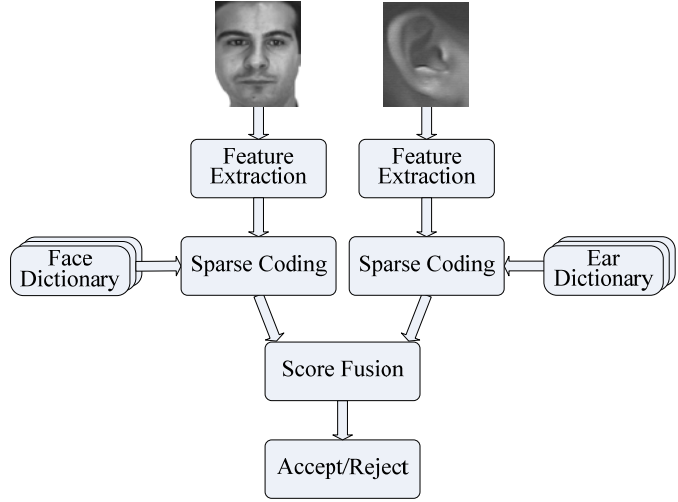

Fig. 2. System model of the proposed SRC-based multimodal verification system using face and ear.

Since the match scores, in terms of SCE or SCR, of face and ear have similar distribution, we hence directly combine them without score normalization for avoiding unexpected possible negative effect of normalization. The proposed SRC-based multimodal verification system shown in Fig. 2 first performs two independent sparse coding procedures for achieving match scores, and then integrates them by using sum-rule fusion. Suppose $A^f = \left[ A_1^f, A_2^f, \cdots, A_c^f \right] \in R^{N \times lc}$ and $A^e = \left[ A_1^e, A_2^e, \cdots, A_c^e \right] \in R^{N \times lc}$ are respectively the face and ear dictionaries, where $A_i = \left[ a_{i,1}, a_{i,2}, \cdots, a_{i,l} \right]$ ($i = 1, 2, \cdots, c$) consists of the training samples of Class $i$. Given the face and ear query samples are $y^f$ and $y^e$, the sparse coding problems can be formalized as follows:

$$\hat{\boldsymbol{\alpha}}^f = \arg\min \left\| \boldsymbol{\alpha}^f \right\|_1 \quad \text{s.t.} \quad \left\| y^f - A^f \boldsymbol{\alpha}^f \right\|_2 < \varepsilon, \tag{6}$$

$$\hat{\boldsymbol{\alpha}}^e = \arg\min \left\| \boldsymbol{\alpha}^e \right\|_1 \quad \text{s.t.} \quad \left\| y^e - A^e \boldsymbol{\alpha}^e \right\|_2 < \varepsilon. \tag{7}$$

The SCE and SCR match scores of face and ear can be calculated by using Eq. (2) and Eq. (4), respectively. For convenience, let $S^f$ and $S^e$ be the sparsity-based match scores. The multimodal match score is obtained by $S = S^f + S^e$. According to the decision metric, the multimodal verification can make decision based on Eq. (3) or Eq. (5).

**2.3 Two major concerns and our solution**

**A. Two major concerns**

As a one-to-many comparison approach and the sparsity constraint, SRC would generally only allow very few classes to get a small SCE, while most classes would get a distinctly large one, see Fig. 1(a). Hence, if the class claimed with valid test sample fails to get a small SCE in the competitive sparse coding, SRC-based verification would very likely give a negative response. In other words, from an identification viewpoint, if a valid test sample cannot be identified at the top few ranks, it seems impossible to be accepted by the SRC-based verification system. That is to say, inferior discrimination/identification capability is unlikely to result in good verification of the SRC system. The SRC system using SCR may also have similar characteristic because generally a high SCE value corresponds to a low SCR value. In Section 3, we will use experiments to confirm that there exists a strong positive correlation between the verification and identification performance of SRC-based biometric systems.

Unexpectedly, we can see in literature, for example, the rank-one recognition rates of SRC and its latest extensions [1-2, 22] are among 94.7%-97.14% on AR face database with 100 subjects [23], and their rank-one recognition rates are among 92.563%-94.781% on USTB III ear database with 79 subjects [15, 24]. Moreover, the practical verification systems usually serve for hundreds or thousands of users. As we know, the increase of dictionary scale would inevitably cause performance degradation for SRC-based identification. Therefore, SRC-based unimodal verification approaches like face and ear may not be able to satisfy verification accuracy requirements in many applications. Furthermore, even if the SRC-based verification can get a very low EER, there might have many genuine scores locate near the distribution center of imposter scores, as shown in Fig. 1. It is impractical to accept them by tuning operating threshold. In other words, the SRC-based verification system using face alone may encounter an evident bottleneck in FRR (see Fig. 8 in Section 3), which may also be unacceptable in some applications.

Another major concern about the SRC-based verification lies in its time consumption. Compared with many common classifiers, the SR-based classification techniques are much more time-consuming due to the high computational complexity of sparse coding. Given the dimensionality $N$ of training sample is fixed, the empirical complexity of commonly used $L_1$-regularized sparse coding methods (such as $l_1-1s$ [25]) to solve Eq. (1) is $O\left((lc)^\vartheta\right)$ with $\vartheta \approx 1.5$. Hence, in the applications with a large number of enrolled subjects, SRC-based verification could be very computationally expensive, or even prohibitive.

**B. Small random dictionary**

The proposed SRC-based multimodal verification using face and ear is able to achieve very promising verification results on two multimodal databases with 50 and 79 subjects, respectively, which will be demonstrated in Section 3. However, on larger scale database, it may still encounter performance reduction in both verification effectiveness and computational cost. For solving these problems, we propose to construct the sparse coding dictionary by randomly selecting training samples of only a small part of enrolled subjects, which is motivated by the considerations below.

The role of non-target classes in SRC-based verification is two-folds. If sufficient training samples are available for each class, it will be possible to faithfully represent the test samples as a linear combination of just those (or some of) training samples from the same class, i.e. $A_i$ [1]. However, face/ear recognition is a typical small-sample-size problem, and $A_i$ is under-complete in general, such that the benefit of sparse representation could not be exploited. Fortunately, face/ear images from different classes are not incoherent but are highly correlated. Their combination $A$ can be an over-complete dictionary, and their collaborative representation of test sample is argued to play a significant role in SRC-based identification [26]. Most importantly, incorporating non-target classes, SRC-based verification provides a competition mechanism to allocate SCE or SCR match score to each class by using $L_1$-norm optimization. Given a genuine test sample, if the claimed class not only achieve favorable match score but also defeat many competitor classes in matching, the possible "accept" decision made by SRC would be more reliable than the conventional verification. Therefore, SCE and SCR match scores are much more meaningful for verification. Most conventional verification methods do not adopt competitor classes mainly owing to the absence of eligible and crafty competition mechanism. Verlinde and Chollet [27] had ever attempted to develop a verification approach with non-target classes by using KNN, but the competition mechanism is too naive to get improved verification performance.

In the SRC-based verification, although the more competitors engaging could indicate the more reliability of the "accept" decision, they may also give rise to wrongful "reject" of genuine test sample. If the claimed class fails to defeat almost all the competitors, with very high possibility, it would just obtain very big SCE or small SCR, and result in "reject" decision. From these viewpoints, a large number of competitor classes may bring about negative effect. In the following, we will illustrate why sparse coding involving excess competitor classes could cause verification degradation.

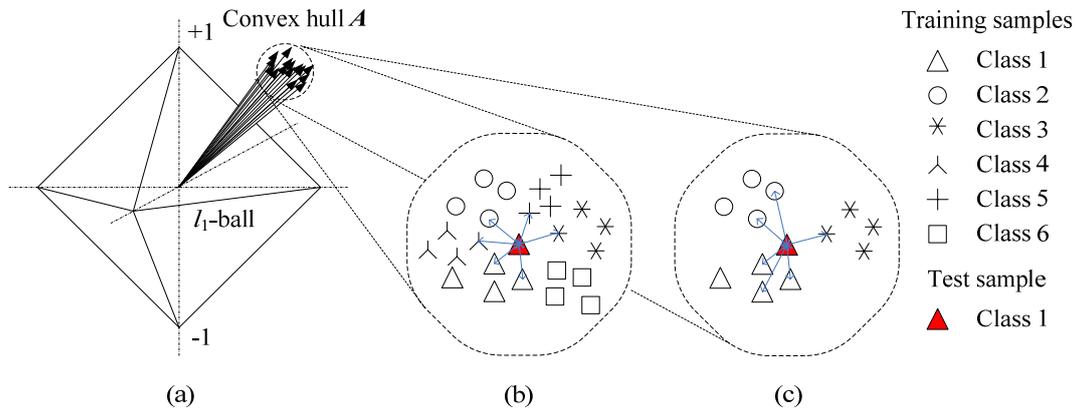

Fig. 3. (a) Convex hull spanned by all columns of $A \in R^{N \times lc}$ on the unit sphere $S^{N-1}$ [28]; (b) - (c) Toy examples illustrating $K$ sparse ($K=6$) $L_0$-norm optimization results on different scale dictionaries.

In face/ear recognition, the columns of $A \in R^{N \times lc}$ are highly correlated: they are all images/features of some face/ear. The convex hull spanned by face/ear images of all subjects in only an extremely tiny portion of the unit sphere $S^{N-1}$, and the training vectors are tightly bundled together as a "bouquet" [28], as shown in Fig. 3 (a). In other words, in the overall convex hull, the distribution interval among classes is very small, and many classes may overlap their neighbors more or less. As for a test sample near/on the boundary, the class claimed may get a big SCE or small SCR, and thus be wrongfully rejected. Moreover, be aware that the convex hull of each class is only an approximation to the true distribution of the face/ear images. In reality, we are not able to collect exhaustive samples of each class to perfectly approach its true distribution. Apart from the limited number of training samples, the noise, expression, and illumination variations, or alignment error, may also aggravate the overlaps among the actual distributions of some classes. Accordingly, more competitors in sparse coding may introduce more distribution overlaps with the claimed class. For simply illustrating the risk of this situation and the benefit of using small random dictionary, we let $N=2$, $c=6$, $l=4$, the convex hull is depicted in Fig. 3 (b). By randomly selecting half of the subjects, a small dictionary consisting of their training samples could be obtained like in Fig. 3 (c). Consider a $K$ sparse $L_0$-norm optimization problem ($K=6$), the number of nonzero entries got by a class could indicate the probability of the test sample belonging to this class, which is similar with SCR. Given a test sample associated to Class 1 near the boundary, in Fig. 3 (c) the genuine match score could be $3/K=0.5$. If using all training samples for coding, the genuine score may become smaller like $2/K=0.33$, as shown in Fig. 3 (b). Such a score is much lower than that achieved in Fig. 3 (c), and hence may lead to a wrongful verification decision.

Additionally, a large number of competitor classes in sparse coding would definitely lead to a heavy computational burden. Selecting a small and suitable number of classes for SRC-based verification is reasonable from viewpoints of both verification accuracy and computational cost. The evener the competitor classes in the convex hull and the less overlap among their distributions, the better performance of SRC-based verification would be. However, since the face/ear feature is generally high dimensional data, it is difficult to measure the overlap degree of different classes. A simple but feasible way is randomly selecting the competitors. The dictionary containing randomly selected competitor classes does not bias any subject. Since the random dictionaries with the same number of competitors would have similar distributions, they may also have similar verification results. In Section 3, we will use many experiments to validate the feasibility of small random dictionary for SRC-based verification, as well as the similar verification performance of the random dictionaries with the same scale. It should be noted that the issue of dictionary scale selection is worthy of studying. Nevertheless, for saving space, we would continue this work in the future.

## 3. Experiments

### 3.1 Databases

Due to the lack of real multimodal database with sufficient samples for each class available, we build up two virtual multimodal databases, namely multimodal database I and II (MD I and MD II), by randomly pairing subjects from face database with subjects from ear database. Georgia Tech (GT) [18] and AR (the first 79 subjects) [23] face databases and USTB III ear database with 79 subjects [24] are selected. Sample images of one person of each database are shown in Fig. 4. For the ear database, the 7 images in red rectangles are used for training, while the other 11 images are used for testing, except two images with very large pose variation. MD I and MD II separately have 50 and 79 virtual subjects, and their compositions are described in details in Table 1. For each virtual subject, we pair 7 face images with 7 ear images to form 7 multimodal samples for training. To obtain more instances for testing, each face probe image is paired with all ear probe images of the same subject. All the face and ear images are normalized to 50×40 pixels.

Table 1. The compositions of MD I and MD II.

| Multimodal database | Sample | | | Face database | Ear database |
|---|---|---|---|---|---|
| MD I | Gallery | 7×50 = 350 | GT | Gallery (7 images/subject randomly selected) | Gallery (50) |
| | Probe | 8×11×50 = 4400 | | Probe (the remainder 8 images/subject) | Probe (50) |
| MD II | Gallery | 7×79 = 563 | AR | Subset 1 (7 images/subject from Session 1) | Gallery |
| | Probe | 7×11×79 = 6083 | | Subset 2 (7 images/subject from Session 2) | Probe |

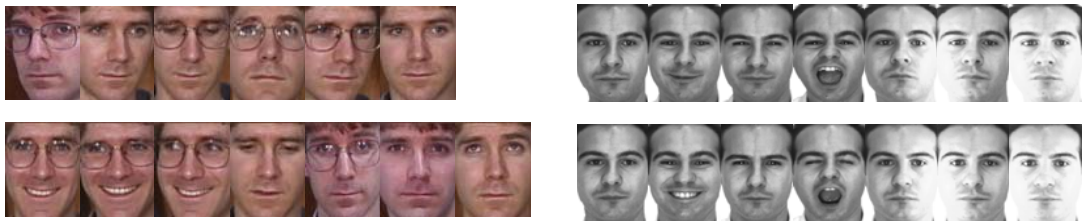

(a) Georgia Tech.  (b) AR (Top: Session 1; Bottom: Session 2).

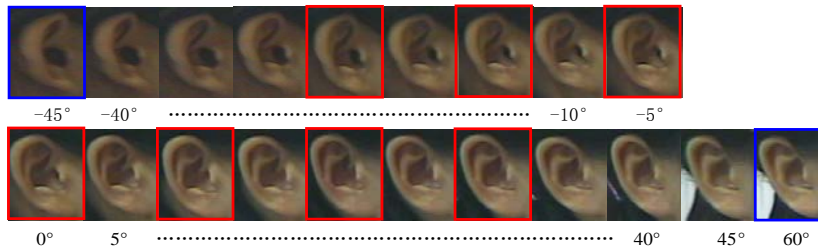

(c) USTB III ear database (Top: Left rotation; Bottom: Right rotation).

Fig. 4. Sample images of one person of the face and ear databases.

### 3.2 Performance evaluation

As for the feature extraction, 2D-DCT is utilized for both modalities since it is fast, general, and without training like subspace methods. The DCT coefficients are scanned in a zigzag manner starting from the top-left corner of the entire transformed image to form a feature vector with 200 dimensions. Feature dimensionality reduction is unnecessary to make the linear equation systems in Eqs. (1), (6) and (7) underdetermined. We use the $L_1$-norm optimization method $l_1$−1s ($\lambda = 0.002$) [25] to solve the sparse coding problems. If no specific instructions, the coding dictionary is composed of training samples of all subjects of the database. All experiments are conducted on Matlab 2010b platform on a desktop with 2.8 GHz dual core CPU and 4 G RAM.

Since the SCE and SCR match scores are derived from the comparison between one-sample and one-class, given a test sample, one genuine score and $c-1$ imposter scores can be obtained. As for the SRC-based multimodal verification, we get 4400 genuine scores and 4400×(50-1)=215600 imposter scores of both SCE and SCR on MD I, while the genuine and imposter scores are respectively 6083 and 6083×(79-1)=474474 on MD II. As for the competing methods using cosine similarity, we select the best match score from the comparisons of a test sample and $l$ training samples of a class, hence the same numbers of genuine and imposter scores with SRC-based methods are available.

Table 2. Biometric verification results in terms of EER (%).

| | | Unimodal | | | Multimodal | | | | |
|---|---|---|---|---|---|---|---|---|---|
| | | Baseline | SRC_scr | SRC_sce | Sum rule | SVM | LLR | SRC_scr | SRC_sce |
| MD I | GT | 18.8 | 4.5 | 4.25 | 12.4 | 5.95 | 6.1 | 0.682 | 0.45 |
| | Ear (50) | 12.8 | 2.34 | 1.64 | | | | | |
| MD II | AR | 16.9 | 2.5 | 1.99 | 11.8 | 6.95 | 6.4 | 0.344 | 0.146 |
| | USTB III | 12.7 | 2.2 | 1.46 | | | | | |

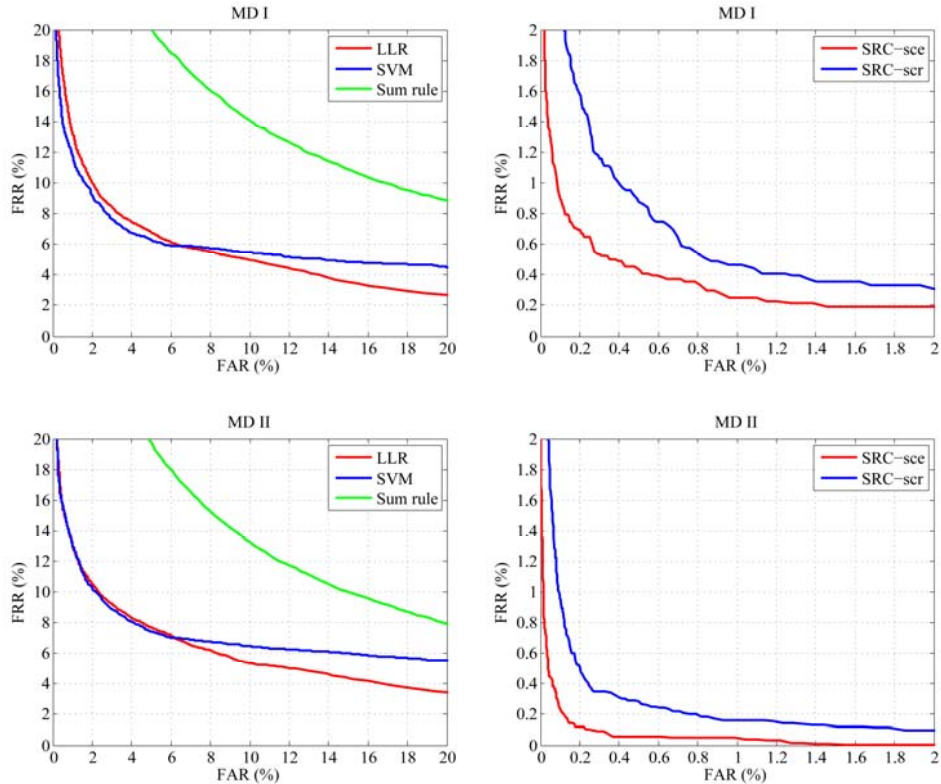

Fig. 5. ROC curves of all competing multimodal methods on MD I and II.

**A. Comparison with previous methods**

Hereafter, we denote the SRC-based verification approaches using SCE and SCR with SRC_sce and SRC_scr, respectively. They are compared with likelihood ratio (LLR) [29], support vector machine (SVM, with RBF kernel: $K(\mathbf{x},\mathbf{y}) = \exp(-15 \times |\mathbf{x}-\mathbf{y}|^2)$) [30], and the standard sum-rule fusion methods, as well as a baseline unimodal method. All the competitors use cosine similarity. The performance is evaluated by false rejection rate (FRR), false acceptance rate (FAR), and their equal error rate (EER). Fig. 5 plots the receiver operating characteristic (ROC) curves of all competing multimodal methods obtained on MD I and II. SRC_sce and SRC_scr are clearly found overwhelmingly superior to the state-of-the-art LLR and SVM. Table 2 summarizes their EER results. We can see that the multimodal SRC_sce gets the best results of 0.45% and 0.146% on MD I and MD II, respectively, while the best results obtained by LLR or SVM among the conventional multimodal methods are only 5.95% and 6.4%. Ear (50) denotes an ear dataset containing the first 50 subjects of USTB III. It can be seen that the face and ear unimodal methods based on SRC also get very impressive results. Either of them gets at least 10% improvement in all cases, compared to the baseline unimodal method. They are even much better than the three well-known multimodal methods.

Overall, these experimental results have validated the great advantage of SRC-based verification over the state-of-the-art methods. Although the face and ear databases are very challenging due to facial expression, illumination, and pose variations, the proposed SRC-based multimodal verification can still achieve very encouraging results. They could potentially be utilized in applications with quite rigorous verification requirement.

Table 3. Biometric identification (Rank-one recognition rate, %) and verification (EER, %) results.

|  |  | Unimodal | | | Multimodal | |
| --- | --- | --- | --- | --- | --- | --- |
|  |  | GT | AR | USTB III | MD I | MD II |
| Identification | | 89.5 | 93.43 | 94.71 | 99.18 | 99.54 |
| Verification | SRC_scr | 4.5 | 2.5 | 2.2 | 0.682 | 0.344 |
|  | SRC_sce | 4.25 | 1.99 | 1.46 | 0.45 | 0.146 |

## B. Correlation with inter-class separability

Here, we use the identification performance as the indicator of inter-class separability. To validate strong correlation between the SRC-based verification and the inter-class separability of samples in dictionary, we summarize their results in Table 3. We can see that without exception in the cases where the higher rank-one identification rate is achieved, both SRC_sce and SRC_scr can get the better verification result in terms of EER. Furthermore, as shown in Fig. 6, both cumulative match characteristic (CMC) curves of SRC-based multimodal identification converge rapidly to 100% accuracy at the first few ranks, which implies that almost all multimodal test samples could lead to very small SCEs or large SCRs of their associated classes. On the contrary, there are still many face or ear samples cannot be identified at rank 5. Correspondingly, SRC-based multimodal methods achieve very good verification results and their superiority to the unimodal methods is significant.

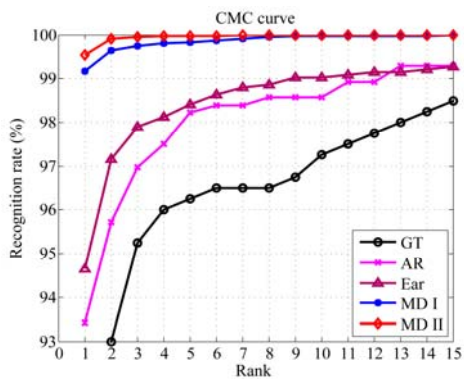 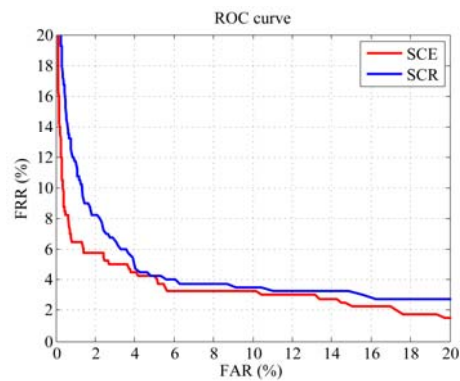

Fig. 6. CMC curves of all SRC-based identification methods.    Fig. 7. SRC-based face verification results on GT face database.

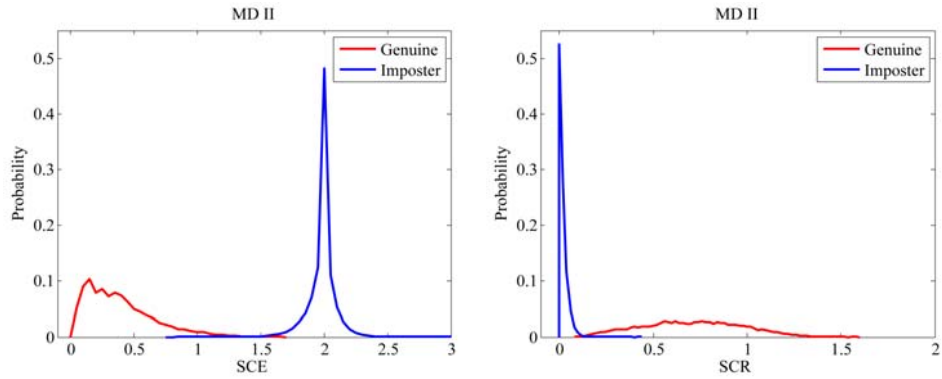

Fig. 8. Distributions of multimodal scores of SCE and SCR on MD II.

Besides, taking GT face database as an example, Fig. 1 have shown that there are many genuine scores locate near the distribution center of imposter scores. Hence, it is impractical to let them pass the verification by tuning operating threshold. That is to say, the SRC-based verification system confronts an evident bottleneck in FRR, as shown by the ROC curves in Fig. 7. Actually, all the unimodal verification approaches using SRC_sce and SRC_scr encounter evident FRR bottleneck, whose ROC curves are not plotted for saving space. On the contrary, the multimodal SRC_sce and SRC_scr nearly have not this drawback, see Fig. 5. Their score distributions on MD II are plotted in Fig. 8.

Overall, when getting inferior identification accuracy, SRC systems generally achieves worse verification results, and meanwhile confronts evident FRR bottleneck. On the other hand, SRC-based system with strong identification capability like the proposed multimodal combination of face and ear is more likely to lead to desirable verification result. Apparently, there exists a positive correlation between the identification and verification performance of SRC-based system. This characteristic could be used as a selection guideline of SRC for verification application using other biometric traits as extensive SRC-based identification reports have already covered almost all the biometric fields.

### C. Small random dictionary

Using small random dictionary would be able to bring a double benefit. The time consumption limitation could be conquered, and SRC applied on a small-scale dictionary should guarantee high discriminative capability. The attendant risk is that whether the derived SCE and SCR match scores are reliable enough to verify an identity. To answer this question, we conduct SRC-based unimodal and multimodal experiments on MD II by using 10 small dictionaries that consist of training samples of 49 randomly selected subjects and the subject claimed for verification. For all multimodal approaches, the number of genuine score is still 6083, while the number of imposter score is 6083×(50-1)=298067.

Table 4. SRC-based identification results (Rank-one recognition rate, %) with different dictionary scales on MD II.

| Scale | AR | USTB III | Multimodal |
|---|---|---|---|
| 79 subjects | 93.43 | 94.71 | 99.54 |
| 50 subjects | 94.16 (±0.4) | 95.62 (±0.2) | 99.61 (±0.04) |

Table 5. SRC-based verification results (EER, %) with different dictionary scales on MD II.

| Scale | AR | | USTB III | | Multimodal | | |
|---|---|---|---|---|---|---|---|
| | SRC_scr | SRC_sce | SRC_scr | SRC_sce | SRC_scr | SRC_sce | Runtime (sec) |
| 79 subjects | 2.5 | 1.99 | 2.2 | 1.46 | 0.344 | 0.146 | 1.416 |
| 50 subjects | 2.8 (±0.15) | 1.96(±0.2) | 2.1(±0.1) | 1.28 (±0.08) | 0.38 (±0.04) | 0.165 (±0.03) | 0.982 |

Table 4 first shows the comparison of the SRC-based identification based on dictionaries with different scales. It can be seen that compared with the dictionary with all 79 subjects, all the 10 small random dictionaries get better identification accuracies. Table 5 outlines the corresponding verification results of SRC_sce and SRC_scr. We can see that small random dictionaries with the same scale achieve similar EERs with only a small variance in both SRC_sce and SRC_scr and in all experiment cases. As for the SRC_sce, all the small dictionaries are obviously better than the big dictionary with all enrolled subjects in ear verification, while half of them excel the latter in the face case. In the multimodal verification, although few small dictionaries get slight accuracy reduction, the majority can be found comparable to or better than the big dictionary. As for the SRC_scr, compared with the big dictionary, all small dictionaries get better or equivalent verification results in the ear case, while some of small dictionaries get comparable verification results in face verification and multimodal verification cases. In the runtime aspect, the multimodal verification experiments show that using small dictionary is obviously more efficient than big dictionary.

In this series of experiments, we can see that compared with the big dictionary, small random dictionaries leading to comparable or even better verification result can generally be found, especially when using SRC_sce. Considering the database scale in the real application would be many times of ours, the SRC-based verification based on big dictionary with all training samples must inevitably suffer from verification degradation and a heavy or even unaffordable computational burden. In such context, selecting a certain small dictionary would be the most effective solution.

## 4. Conclusion

In this paper, we have first gave an insight into SRC-based biometric verification by studying two sparsity-based metrics (SCE and SCR) on two biometric traits (face and ear) and their multimodal combination with sum-rule fusion at match score level. In our experiments, both multimodal verification methods using SCE and SCR are significantly superior to the state-of-the-art likelihood ratio (LLR) and support vector machine (SVM) using cosine similarity. Even the face and ear SRC-based unimodal verification methods are much better than these conventional multimodal methods. Secondly, it can be seen that with the same SRC verification frameworks, face and ear unimodal methods are tremendously inferior to their multimodal combination, and both confront evident FRR bottleneck on all databases while the latter does not have this phenomenon in our experiments. These drawbacks can be attributed to the relatively inferior inter-class separability in unimodal scenarios. As we can see that there exists an obvious positive correlation between the identification and verification performance of SRC systems. Such a strong correlation could be seen as a selection guideline of SRC for verification applications based on the extensive SRC-based biometric identification reports in literature. More importantly, for simultaneously settling SRC's limitations in verification effectiveness and computational efficiency in large-scale application scenarios, we have proposed to exploit small random dictionary for sparsity-based score computation. Lots of SRC-based verification experiments on face, ear, and their combination have confirmed that using small random dictionary is feasible in terms of both effectiveness and efficiency. Besides, it is observed that random dictionaries with the same scale generally bring about very similar SRC-based verification performance. In the future work, two issues should be studied in deep, including what scales of the random dictionary are favorable, and how to optimally select or optimize the random dictionary.

## 5. Acknowledgement

This work is supported by NSFC under Grants 61173182, 61179071, 61303126, 61411130133, by funding from Sichuan Province under Grants 2014HH0048, by funding from Xihua University under Grants 20141120, as well as by the Young foundation of School of Computer and Software Engineering.